\documentclass{article}

\usepackage{arxiv}

\usepackage[utf8]{inputenc} 
\usepackage[T1]{fontenc}    
\usepackage{hyperref}       
\usepackage{url}            
\usepackage{booktabs}       
\usepackage{amsfonts}       
\usepackage{nicefrac}       
\usepackage{microtype}      
\usepackage{lipsum}
\usepackage{graphicx}
\graphicspath{ {./images/} }

\usepackage{amsmath,epstopdf,flushend,amssymb} 
\usepackage[figuresright]{rotating}
\usepackage{multirow,multicol,mathtools,comment}
\usepackage{xcolor}
\usepackage{pifont}
\newcommand{\cmark}{\ding{52}}%
\newcommand{\xmark}{\ding{56}}%

\usepackage{subfigure}

\title{Seen to Unseen: When Fuzzy Inference System Predicts IoT Device Positioning Labels That Had Not Appeared in Training Phase}
  
\author{
    Han Xu \\
  Institute of Microelectronics, Chinese Academy of Sciences\\
  University of Chinese Academy of Sciences\\
  Beijing 100029, China\\
  \texttt{ann.gong.qifeng@gmail.com} \\
  \And
    Zheming Zuo\\
  Department of Computer Science\\
  Durham University\\
  Durham DH1 3LE, UK\\
  \texttt{zheming.zuo@durham.ac.uk} \\
  \And
  \And
  Jie Li $\qquad\qquad\qquad\qquad$ Victor Chang\thanks{Corresponding author.}\\
  Cybersecurity, Information Systems and AI Research Group\\
  School of Computing, Engineering \& Digital Technologies\\
  Teesside University\\
  Middlesbrough TS3 6DR, UK\\
  \texttt{\{jie.li,v.chang\}@tees.ac.uk} \\
}

\begin{document}
\maketitle
\begin{abstract}
Situating at the core of Artificial Intelligence (AI), Machine Learning (ML), and more specifically, Deep Learning (DL) have embraced great success in the past two decades. However, unseen class label prediction is far less explored due to missing classes being invisible in training ML or DL models. In this work, we propose a fuzzy inference system to cope with such a challenge by adopting TSK+ fuzzy inference engine in conjunction with the Curvature-based Feature Selection (CFS) method. The practical feasibility of our system has been evaluated by predicting the positioning labels of networking devices within the realm of the Internet of Things (IoT). Competitive prediction performance confirms the efficiency and efficacy of our system, especially when a large number of continuous class labels are unseen during the model training stage.
\end{abstract}


	\section{\uppercase{Introduction}}\label{sec:introduction}

Indoor Positioning System (IPS) is a tracking system that usually uses a set of network devices to locate people or objects within a building, or a particular room \cite{kim2021review} where GPS would fail entirely \cite{9500208} or lack of high accuracy \cite{suroso2021random}. An indoor tracking system is one of the most helpful features of a smart building \cite{9568311} or smart environment \cite{9389926} that can usually be achieved by working with Internet of Things (IoT) \cite{9261582,vcchang_iot_pos} techniques, \emph{e.g.} WiFi and Bluetooth technologies. Currently, such IoT-based indoor tracking systems have been applied in several areas that lead to \textbf{practical} \textit{benefits}. For instance, \textcolor{blue}{an indoor navigation system is presented in} \cite{khanh2020wi} to track the location of a self-driving cart, thus helping to figure out the shortest path in a smart indoor environment in real-time. A hospital equipment tracking system for the well-being and safety of the patients is also discussed in \cite{curran2011evaluation}. Those systems normally use modulated WiFi/Bluetooth transmission signals to determine whether the service exists by adopting either the triangulation method or Machine Learning (ML) \cite{jordan2015machine} algorithms. Triangulation, \emph{i.e.} trilateration, is a well-established method, which calculates the distance between the tracking object and WiFi/Bluetooth Access Points (APs) by measuring the Received Signal Strength Indicator (RSSI) \cite{8552138}. Nonetheless, the aforementioned methods usually require to use of beacon signals for the precise estimation of distance. A large number of beacons can cause interference with each other. 

\begin{figure}[!ht]
	\centerline{\includegraphics[width=0.8\textwidth]{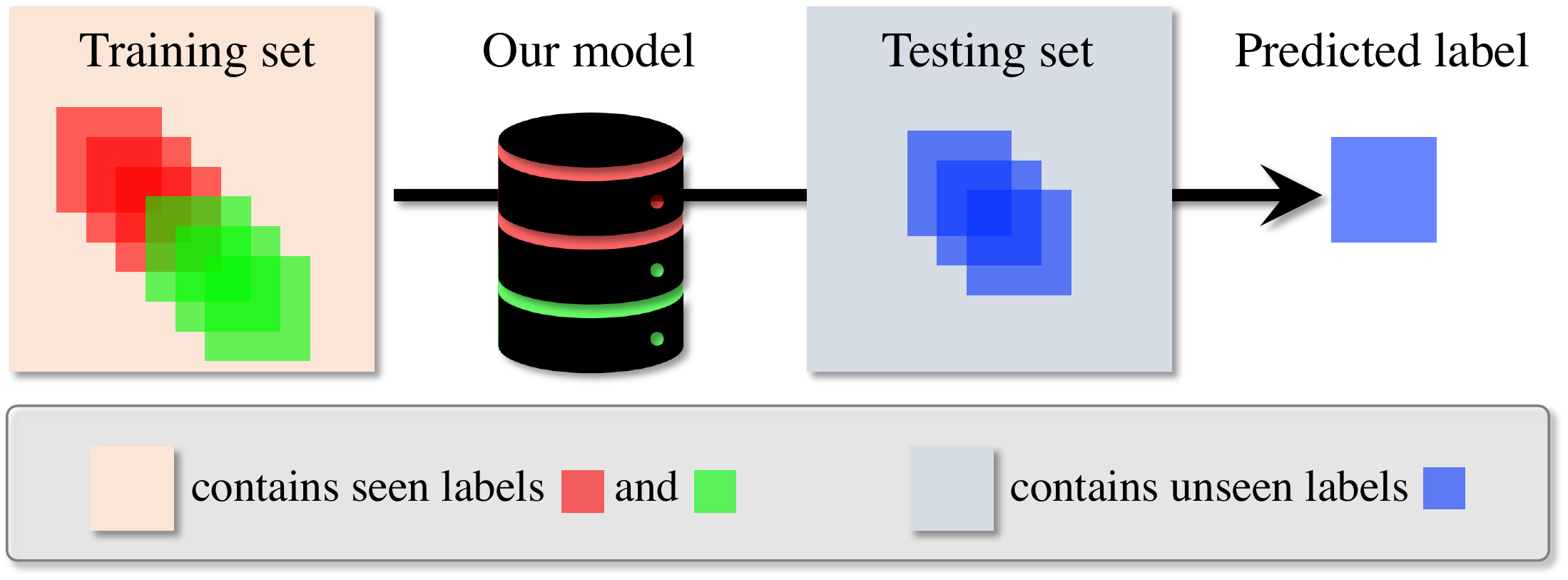}}
	\caption{The motivation of this work is to build a model which predicts the IoT device positioning labels that were not appeared during the training phase.}
	\label{fig:mov}
\end{figure}

In contrast, ML and Deep Learning (DL) \cite{lecun2015deep} were employed to transfer an indoor tracking problem into a classification problem. In recent year, ML and DL have achieved great success with a wide spectrum of applications including action recognition in videos \cite{8919994}, face detection in low-light conditions \cite{9049390}, image denoising in low-light and noisy scenes \cite{Zuo_IdeaNet_2022_WACV} etc. In the context of IPS, the RSSIs values are collected from a set of pre-deployed beacon devices to form the training dataset at each of the known locations. From there, a system model would be trained based on all the given training instances. Given an input, which contains multiple RSSI values, ML predicts the corresponding location in line with the trained model \cite{nessa2020survey}. However, none of those models are capable of predicting the label of a testing instance if that specific ground-truth label is not visible to the model during the training phase. That is, as shown in Fig. \ref{fig:mov}, \emph{e.g.}, a model is trained on the instances with class labels of $c_{\text{\textcolor{red}{red}}}$ and $c_{\text{\textcolor{green}{green}}}$ (`seen labels'), and testing on the ones with ground-truth label of $c_{\text{\textcolor{blue}{blue}}}$(`unseen label'). 

Fuzzy inference systems \cite{YangZuoInTech2017} are usually built upon fuzzy logic theory to map the inputs of the system to desired outputs. Typically, a fuzzy inference system includes a rule base and an inference engine. In terms of the inference engine, Mamdani \cite{mamdani1977application} and TSK\cite{TSK1985} are the two most common choices. In the conventional settings, the fuzzy inference approaches are only compatible with dense rule bases, and each of which covers the entire input domain. Fuzzy interpolation techniques \cite{koczy1993approximate,Jie2017,li2018interval,li2018extended} were originally proposed to ease the requirement of the dense rule base, which allows the fuzzy inference to be performed over a sparse rule base \cite{li2016experience}. That is, the conclusion still could be drawn via interpolation in the case that a given input (\emph{i.e.} observation) does not overlap with any rule antecedent. Thereby, the task of `unseen' label prediction could be converted to a sparse rule base problem where the fuzzy interpolation technique is adopted. In this study, TSK+ \cite{li2018extended}, as one of the TSK-style fuzzy interpolation approaches, is utilized to deal with the indoor tracking system with the following major contributions:
\begin{itemize}
    \item Proposed a fuzzy inference system by adopting TSK+ fuzzy interpolation approach to address the indoor location tracking problem with invisible classes during the training phase;  
    \item Deployed a Curvature-based Feature Selection (CFS) method \cite{zuo2021curvature} to reduce the space and time complexity of the proposed fuzzy inference system;
    \item Evaluated the performance of the proposed fuzzy inference system using a real-world IoT device positioning dataset with competitive prediction performance yielded.
\end{itemize}

\section{\uppercase{Background}}\label{sec:bg}
The extended Takagi–Sugeno–Kang (TSK+), as a fuzzy interpolation method \cite{YangZuoInTech2017}, is capable of conducting fuzzy inferences on a sparse TSK rule base, which is introduced in the first half of this section. This is followed by a brief review of a feature selection method, curvature-based feature selection, adopted in our system.

\subsection{TSK+ Fuzzy Interpolation}\label{sec:tsk+}
Conventional TSK system generates a crisp inference result (\emph{i.e.} output) from a given input by calculating the weighted average over the sub-consequences of all fired rules~\cite{TSK1985}. Obviously, the firing strength of all rules is valued as zero in the case that a given input does not overlap with any rule antecedent, and thus, no consequence can be derived. TSK-interpolation has addressed such issue~\cite{Jie2017}. Suppose that a sparse TSK rule base $\mathcal{R}$ is comprised of $n$ rules:
\begin{equation}\label{TSKRB}
\begin{aligned}
\mathcal{R}_1:\ & \textbf{IF}\ x_1\text{ is } \mathcal{A}_1^1 \text{ and } x_2\text{ is } \mathcal{A}_2^1 \text{ and } \cdots  \text{ and } x_m\text{ is } \mathcal{A}_m^1  \\
& \textbf{THEN}\  \gamma=\Gamma_1,	\\
&...\ ...\\
\mathcal{R}_i:\ & \textbf{IF}\ x_1\text{ is } \mathcal{A}_1^i \text{ and } x_2\text{ is } \mathcal{A}_2^i \text{ and } \cdots  \text{ and } x_m\text{ is } \mathcal{A}_m^i  \\
& \textbf{THEN}\  \gamma=\Gamma_i,	\\
&...\ ...\\
\mathcal{R}_n:\ & \textbf{IF}\ x_1\text{ is } \mathcal{A}_1^n \text{ and } x_2\text{ is } \mathcal{A}_2^n \text{ and } \cdots  \text{ and } x_m\text{ is } \mathcal{A}_m^n  \\
& \textbf{THEN}\  \gamma=\Gamma_n,	\\
\end{aligned}
\end{equation}
where $n$ denotes the the number of rules (\emph{i.e.} size of the rule base); $\mathcal{A}_k^i$ denotes a normal and convex fuzzy set s.t. $i\in[1,\cdots, n]$ and $k\in[1, \cdots , m]$, and $\gamma$ is the consequent of a rule base. In addition, we utilize triangular membership functions for simplicity, and therefore $\mathcal{A}_{k}^i$ is conveniently denoted as ($a_{k1}^i, a_{k2}^i, a_{k3}^i$), where ($a_{k1}^i, a_{k3}^i$) represents the support of the fuzzy set and $a_{k2}^i$ denotes the normal point. Regardless whether an input $\mathcal{I}=(\mathcal{A}_1^*,\mathcal{A}_2^*,\cdots,\mathcal{A}_m^*)$ overlaps with any rule antecedents, the desired crisp output can be calculated via the following three steps.\\

\noindent \textbf{Step 1: Matching Degree Determination.} $\quad$ The matching degrees $\mathcal{S}(\mathcal{A}_1^*, \mathcal{A}_1^i$), $\mathcal{S}(\mathcal{A}_2^*, \mathcal{A}_2^i)$, $\cdots$, and $\mathcal{S}(\mathcal{A}_m^*, \mathcal{A}_m^i)$ between the inputs ($\mathcal{A}_1^*,\ \mathcal{A}_2^*,\cdots, \mathcal{A}_m^*$) and corresponding rule antecedents ($\mathcal{A}_1^i,\ \mathcal{A}_2^i,\cdots, \mathcal{A}_m^i$) for each rule $\mathcal{R}_i$ ($i\in{1,2, \cdots, n}$) can be determined by:
\begin{equation}\label{eq:similarityequation}
\begin{split}
\mathcal{S}(\mathcal{A}_{k}^i,\mathcal{A}_k^*) =  \Bigg( 1- \frac{\sum_{j=1}^{3} |a_{kj}^i-a^*_{kj}|}{3} \Bigg) \cdot \Bigg(\mathcal{D}\Bigg), 
\end{split}
\end{equation}
where $\mathcal{D}$, namely distance factor, is a function of the distance between the two concerned fuzzy sets. $\mathcal{D}$ is, in turn, computed as:
\begin{equation}\label{dfcalculation}
\mathcal{D} = 1 - \frac{1}{1+e^{-h d +\omega}}\ ,
\end{equation}
where $h$, s.t. $h > 0$, is a sensitivity factor, $d$ represents the distance between the two fuzzy sets, and $\omega$ is a constant. A smaller value of $h$ leads to the similarity degree which is more sensitive to the distance of two fuzzy sets and vice versa. 

\noindent \textbf{Step 2: Firing Degree Calculation.} $\quad$ Compute the rule-wise firing degree ($\theta_i$) by integrating the matching degrees of its antecedents and the known input values:
\begin{equation}\label{alphaI}
\theta_i = \mathcal{S}(\mathcal{A}_1^*, \mathcal{A}_1^i) \wedge \mathcal{S}(\mathcal{A}_2^*, \mathcal{A}_2^i) \wedge \cdots \wedge \mathcal{S}(\mathcal{A}_m^*, \mathcal{A}_m^i)\ ,
\end{equation}
in which $\wedge$ represents the t-norm that is practically implemented as a minimum operator. 

\noindent \textbf{Step 3: Sub-consequence integration.} $\quad$ Generate the final output ($\gamma$) via the integration of the sub-consequences from all the rules by:
\begin{equation}\label{eq:resultTSK}
\gamma=\frac{\sum_{i=1}^{n}\theta_i \cdot \Gamma_n\ } {\ \sum_{i=1}^{n}\theta_i\ } .
\end{equation}

\subsection{Curvature-based Feature Selection} \label{sec:cfs}
Feature Selection (FS) aims to select a subset of the most relevant attributes for the use of model construction from the given dataset \cite{zuo2021curvature}. In particular, FS methods identify the feature-wise importance for a given problem, thus helping select the most relevant (or discriminative) features. Curvature-based FS (CFS) method \cite{zuo2021curvature} was developed to rank the weights (\emph{importance}) of features by calculating the averaged Menger Curvature \cite{leger1999menger} of each feature, which can be summarized into the following three steps:

\noindent \textbf{Step 1: 2-D Data Re-construction.} $\quad$ Given a high-dimensional dataset $\mathcal{M}$, which contains $n$ attributes, donates as $\mathcal{F}_i$ ($ 1\leqslant i \leqslant n$), $\mathcal{M}$ can be divided into $n$ 2D-panels, and each of which can be represented as $\mathcal{P}_{(\mathcal{F}_i,y)}$, where $y$ is the index of the data instance. 

\noindent \textbf{Step 2: Feature Weights Calculation.} $\quad$ For each 2-D panel $\mathcal{P}_{(\mathcal{F}_i,y)}$, the Menger Curvature approach is employed to calculate the averaged curvature value of the feature $\mathcal{F}_i$. For a given 2-D panel ($\mathcal{P}_{(\mathcal{F}_i,y)}$) that includes $m$ data instances, the Menger Curvature value ($\mathcal{C}_{m_j}^i$) of data point $m_j$ (s.t. $j\in[2,m-1]$) can be expressed by: 
\begin{equation}\label{eq:MCvalue}
	\mathcal{C}(m_{j-1}, m, m_{j+1}) = \frac{1}{R} = \frac{2\textrm{sin}(\varphi)}{\|m_{J-1},m_{j+1}\|},
\end{equation}
Thereby, the mean of $\mathcal{C}$ for $\mathcal{F}_i$, denoted as $\widehat{\mathcal{C}_{\mathcal{F}_i}}$, is calculated as:
	\begin{equation}
		\widehat{\mathcal{C}_{\mathcal{F}_i}} = \frac{1}{m-2}\sum_{j=2}^{m-1} \mathcal{C}_{m_j}^i,
	\end{equation}
where $\mathcal{C}_{m_j}^i$ represents the curvature value of the $m_j^{th}$ data point in feature $\mathcal{F}_i$. $\widehat{\mathcal{C}_{\mathcal{F}_i}}$ indicates the corresponding weight of $\mathcal{F}_i$. Furthermore, the higher value of $\widehat{\mathcal{C}_{\mathcal{F}_i}}$, the importance of the feature $\mathcal{F}_i$ for $\mathcal{M}$, and vice versa. 

\noindent \textbf{Step 3 Feature Ranking and Selection.} $\quad$ The features' weight can be ranked by adopting a conventional ordinal ranking method. That is, features with $\widehat{\mathcal{C}_{\mathcal{F}}}$ greater than the given threshold $\epsilon$ are chosen for further data modelling. (The determination of the threshold $\epsilon$ would be based on the given situation, which will remain as the future work.) 

\begin{figure*}[]
	\centerline{\includegraphics[width=0.99\textwidth]{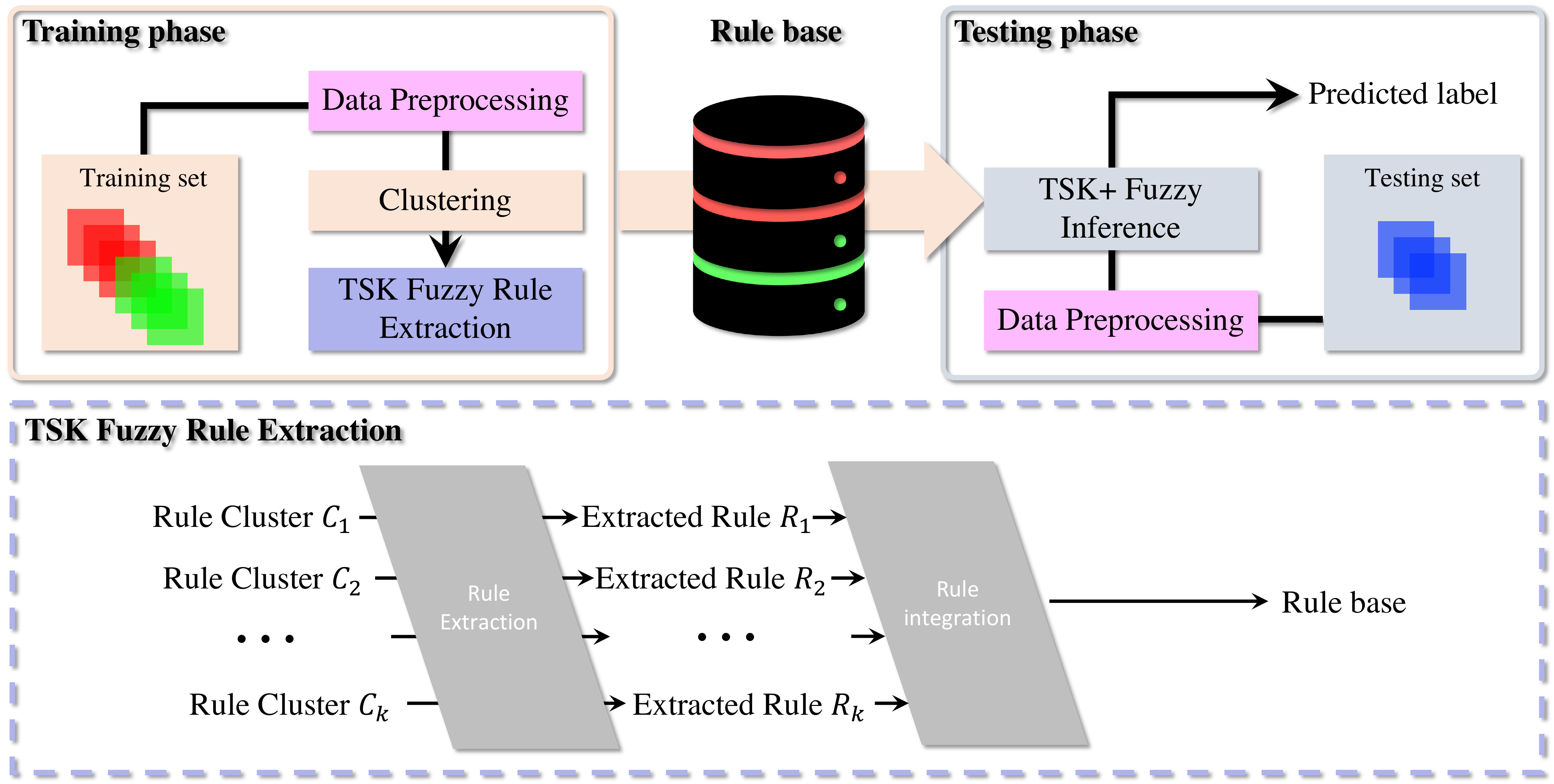}}
	\caption{The pipeline of the proposed system. Note that the `Data Preprocessing' component contains the feature normalization and selection.}
	\label{fig:pip}
\end{figure*}

\section{\uppercase{Proposed System}}\label{sec:proposedsystem}
The proposed fuzzy-based indoor tracking system is presented in this section, which adopts with the introduction of using signal strength between WiFi or Bluetooth signal receiver and different Reference Points (RP), such as WiFi APs and Bluetooth beacon, to predict the position of the device in a WiFi/Bluetooth enabled environment. Our system contains two major components, including a TSK rule base generation (\emph{i.e.} training phase) and a psition prediction (termed as testing phase). Figure \ref{fig:pip} illustrates the pipeline of the proposed system. Particularly, our system first constructs a fuzzy rule base from a visible training dataset by employing the feature selection (\emph{i.e.} CFS) and clustering (\emph{i.e.} $k$-Means) techniques. From there, given a testing data instance, the corresponding position/location would be predicted by adopting TSK+ fuzzy inference approach \cite{li2018extended}. The proposed indoor system is capable of predicting unseen position/label. In other words, the position information had not appeared in the training dataset at all. This is achievable mainly due to the fact that the TSK+ method performs inference using a sparse rule base.

\subsection{TSK Fuzzy Rule Base Generation}\label{sec:rbgen}
A data-driven TSK-style rule base generation approach, which was proposed in \cite{li2018extended} and illustrated in the bottom part of Figure \ref{fig:pip}, is applied in this work to generate a rule base for TSK+ approach. Particularly, given a sparse dataset $\mathcal{T}$, that lacks some locations information, the system first applies the feature selection method CFS where all the features are normalized using the min-max method \cite{zuo2018grooming,8858838}, introduced in Section \ref{sec:cfs}, to reduce the dimensionality of the given dataset. This is followed by the $k$-Means clustering algorithm, which is applied to group all the data instances into a number of clusters. From there, each cluster is expressed as a TSK rule. Lastly, our system combines all the extracted rules and generates the final TSK rule-based. The processes of clustering and fuzzy rule extraction are detailed below:

\begin{itemize}
    \item \textbf{Optimal Number of Clusters Determination.} $\quad$ The Elbow method is carried out in this work to determine the optimal $k$ in $k$-Means clustering algorithm, as a faster and effective method \cite{kodinariya2013review}. In particular, this method determines the number of clusters by adding another cluster that does not lead to a much better modeling result. For example, given a problem, the relationship between performance improvement and the number of clusters is shown in Figure \ref{fig:noK}. The value of $k$ can be obtained as 4, which is determined as a turning point (\emph{i.e.} Elbow point). 

\begin{figure}[!ht]
	\centerline{\includegraphics[width=0.8\textwidth]{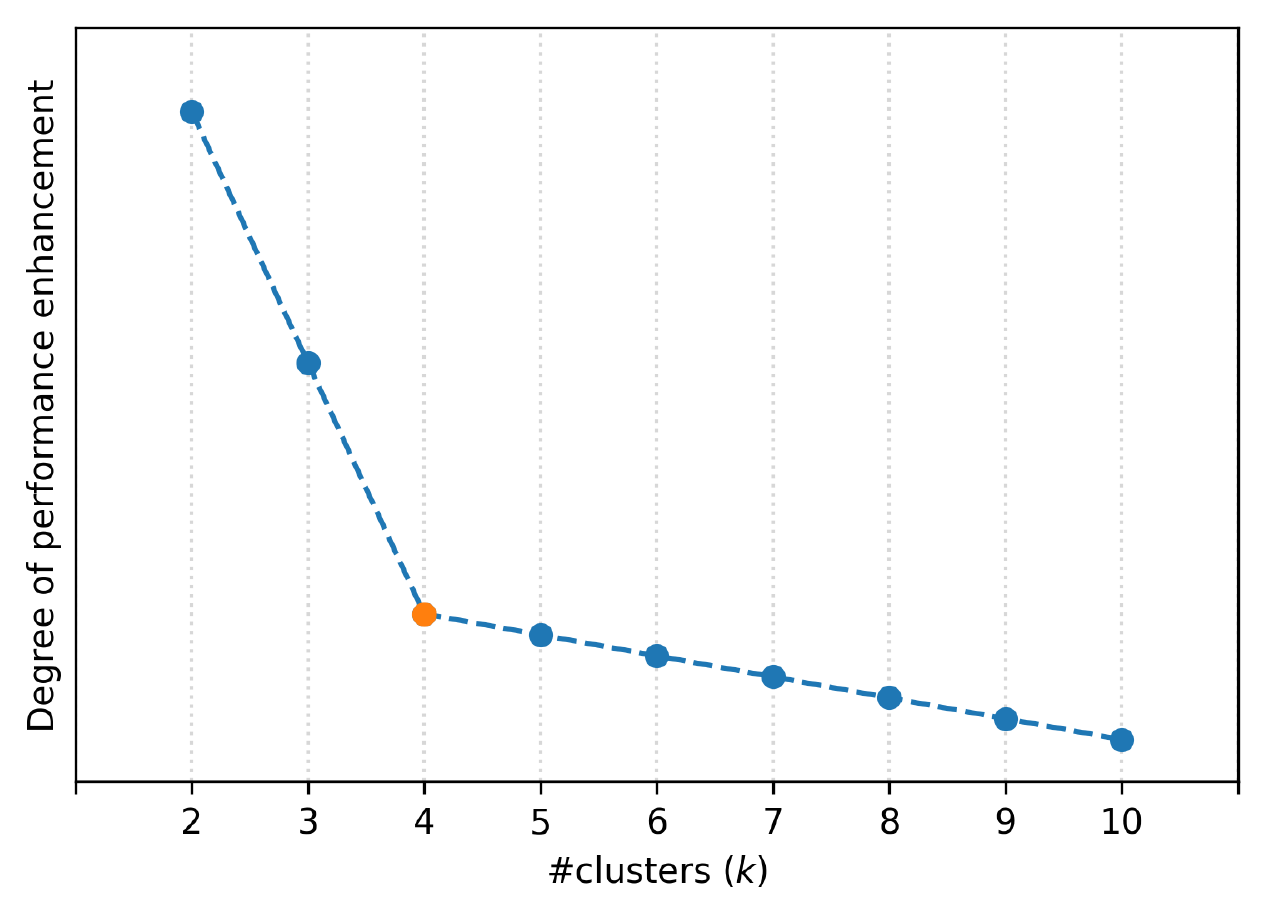}}
	\caption{Examination of $k$ by the Elbow method.}
	\label{fig:noK}
\end{figure}
    
    \item \textbf{Fuzzy rule extraction.} $\quad$ Each obtained cluster is used to form one TSK fuzzy rule. In this work, triangular membership functions are utilized. Given a cluster $\mathcal{C}_r, (1 \leq r \leq k)$, which is determined from the above step, a TSK fuzzy rule $R_r$ can be extracted as:
\begin{equation}\label{eq:tsk}
\begin{aligned}
R_r:\ 	&	\textbf{IF}\ x_1\ \text{is}\ A_{1r}\text{ and } \cdots \text{ and } x_n\ is\ A_{nr}\ \\
&	\textbf{THEN}\  y = \mathcal{L} .
\end{aligned}
\end{equation}
where $A_{sr}$ ($1\leq s \leq n$) are rule antecedents, which are triangular fuzzy sets, represented as $(a^1_{sr}, a^2_{sr}, a^3_{sr})$. Suppose that $C_r$ has $p_r$ elements, \emph{i.e.} $C_r = \{x_r^1,$ $ x_r^2, \cdots,x_r^{p_r}\}$, the core of the triangular fuzzy set is set as the cluster centroid, that is $a^2_{sr} = \sum_{q=1}^{p_r} x_r^q/p_r$; and the support of the fuzzy set is set as the span of the cluster, \emph{i.e.} $(a^1_{sr}, a^3_{sr}) = (\text{min}\{x^1_r, x^2_r, \cdots, x_r^{p_r}\},$ $\text{max}\{x^1_r, x^2_r, \cdots, x^{p_r}_r\})$. $\mathcal{L}$ is rule conclusion, which are discrete integer numbers to represent the corresponding class labels. 
\end{itemize}

\subsection{Unseen Location Prediction by TSK+}\label{sec:prediction}
Thanks to the characteristics of the TSK+ inference approach, which allows the fuzzy inference to be performed over a sparse rule base. Naturally, TSK+ method, as briefed in Section \ref{sec:tsk+}, is readily utilized as a classifier to perform inference. Given a testing data instance $O$ that contains several WiFi or Bluetooth signal strength information, which collected in the same environment as the training phase, but corresponding location label has not appeared in the training data. From there, the TSK+ approach first calculates the matching degrees between the given inputs $O$ and the rule antecedents of each existing rule using Eq. (\ref{eq:similarityequation}). Then, the prediction results of the location information are produced from Eq. (\ref{eq:resultTSK}).

Note that, although a number of machine learning algorithms have been adopted to solve the IoT sensor-based indoor location tracking problems, such as Artificial Neural Network (ANN) \cite{9138275}, $k$-nearest Neighbour ($k$NN) \cite{9138275}, Decision Tree (DT) \cite{alhajri2018classification} and Support Vector Machine (SVM) \cite{alhajri2018classification}, all those systems require location information to be learned in the training phase. Compared with such existing systems, which would not be able to deal with unseen location information, the proposed system can still make a decision for unseen labels. 

\section{\uppercase{Experiments}}\label{sec:experiments}
In this section, we firstly justify the practical feasibility of the proposed system in predicting the unseen label using a dataset collected for indoor location tracking within the IoT sector. Then, we further deploy our system with a fast feature selection method to improve prediction performance.

\subsection{Dataset and Experimental Design}
We employ the \textbf{Miskolc IIS} dataset \cite{toth2016miskolcIIS} in this study, which was collected for hybrid indoor positioning. Generally, it contains a total of 
\begin{figure}[!ht]
	\centerline{\includegraphics[width=0.8\textwidth]{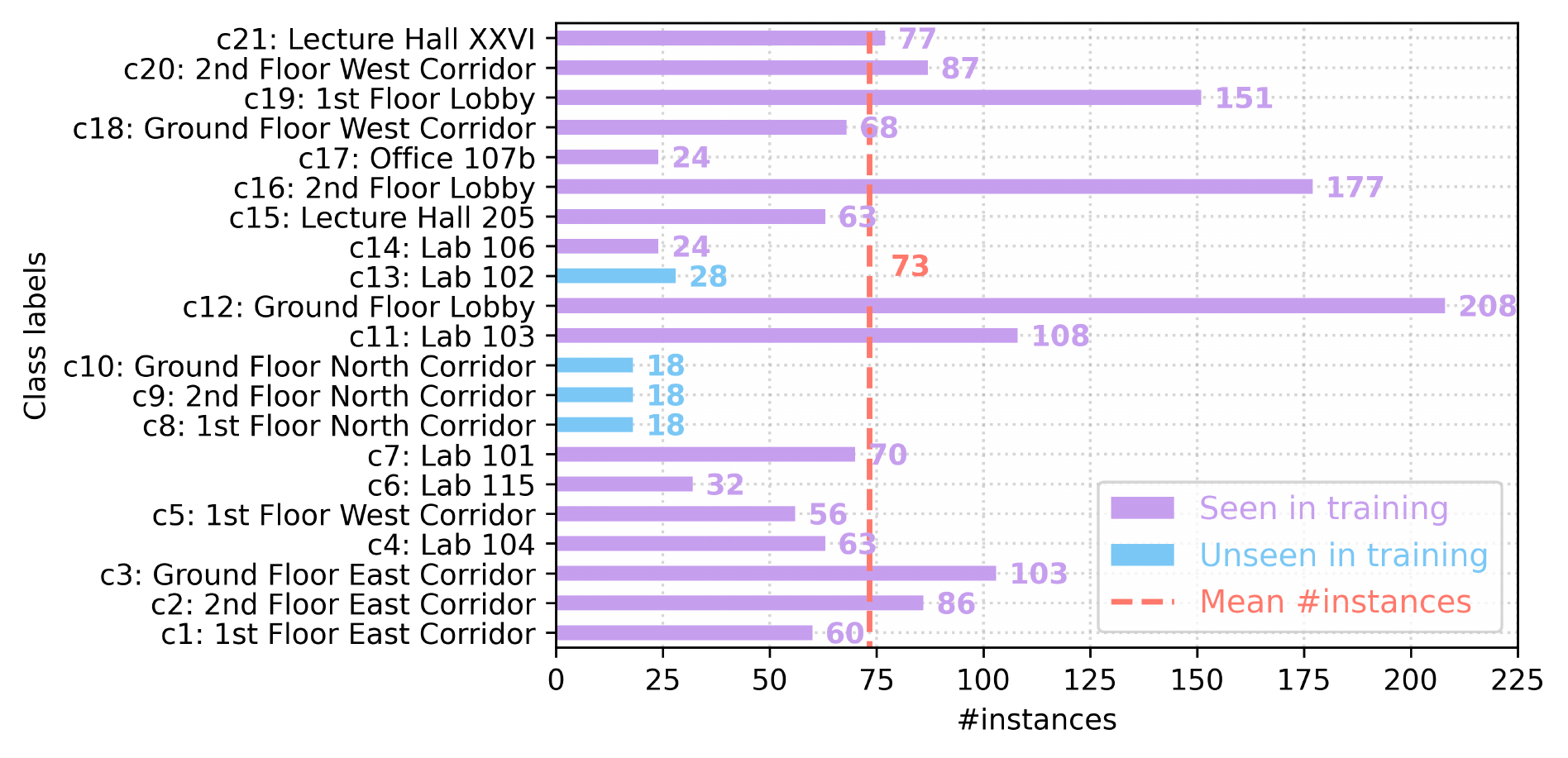}}
	\caption{Class label distribution of the Miskolc IIS dataset.}
	\label{fig:dataset_attr}
\end{figure}

\noindent 1,540 instances, each one including 65 attributes. We use 29 attributes in this study because the rest of attributes, e.g., the actual location coordinates, are irrelevant to this experimentation. In addition, all the instances are categorized into 21 classes. this dataset is an unbalanced dataset, as the most of classes have more than 200 data instances and `c8', `c9' and `c10' have only 18 data instances. Concretely, we visualize the distribution of the attributes in Figure \ref{fig:dataset_attr}. 

\begin{figure}[!ht]
	\centerline{\includegraphics[width=0.8\textwidth]{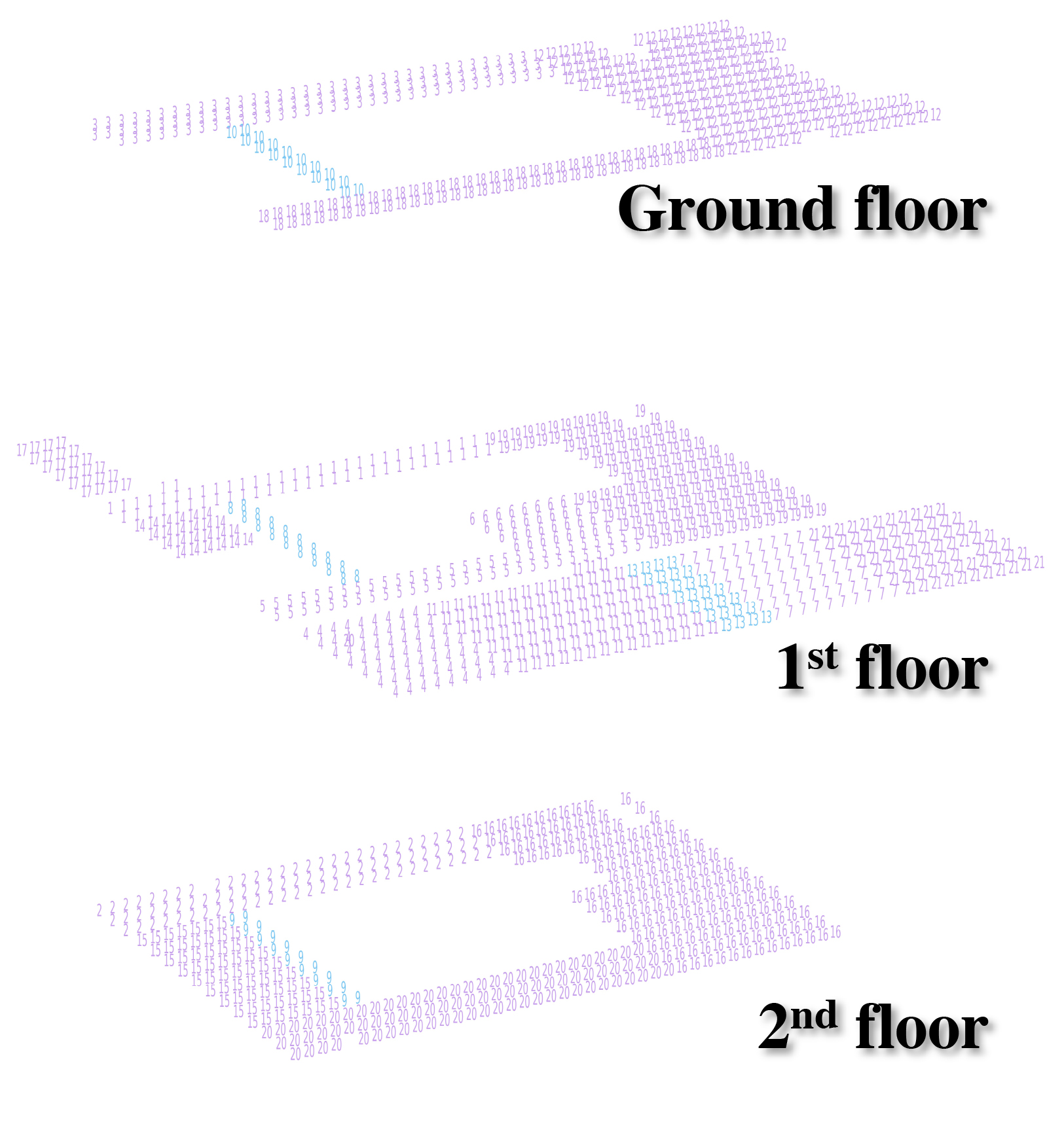}}
	\caption{Visualisation of the floor plan where 21 color numbers denote all the room located in all layers. Color codes are consistent with those adopted in Figure \ref{fig:dataset_attr}.}
	\label{fig:floor_map}
\end{figure}

On this basis, two experimental scenarios are designed: 1) continuous unseen label prediction (\emph{i.e.} instances from classes of `c8', `c9' and `c10' will not appear during the model training phase, due to containing less data instances); 2) discrete prediction of the unseen label (\emph{i.e.} we only make instances of class `c13' (28 instances) available in the testing phase). Thus, as depicted in Figure \ref{fig:floor_map}, data instances that are seen and unseen during the training phase are visualized in conjunction with the actual three-dimensional positions within a 3-layer building of \emph{Miskolc IIS} dataset.

\begin{figure*}[!ht]
	\centerline{\includegraphics[width=0.98\textwidth]{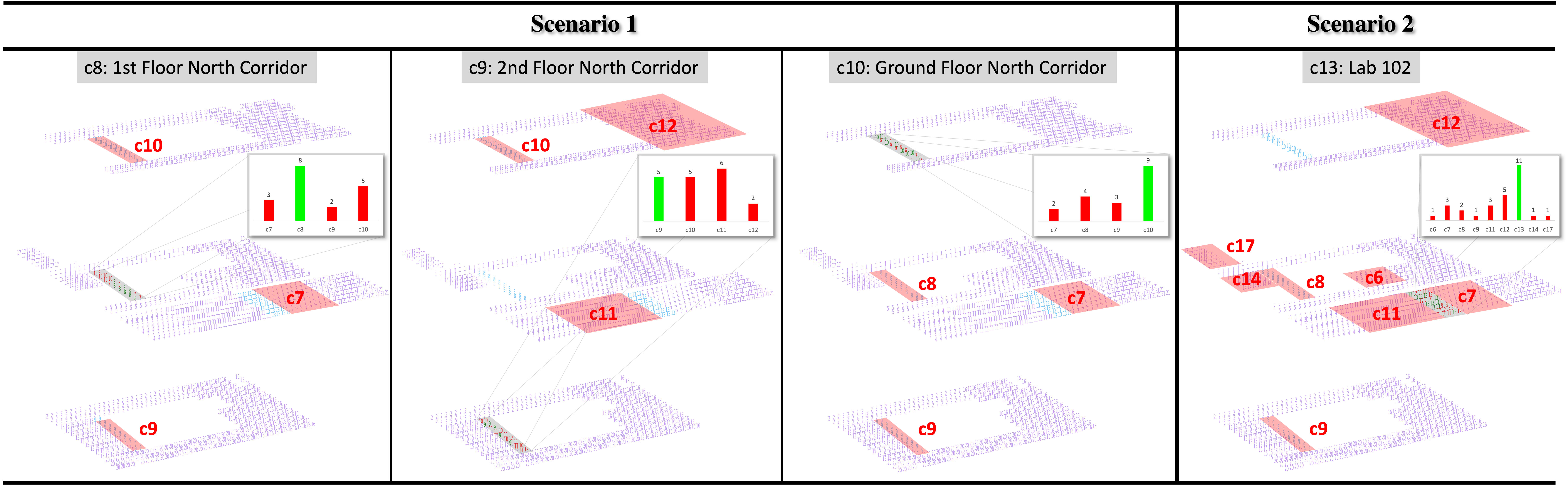}}
	\caption{Experimental results for unseen class label prediction without using CFS. Best viewed in color and zoomed mode.}
	\label{fig:exp_results}
\end{figure*}

\subsection{Feature Importance Ranking}
\begin{figure}[!ht]
	\centerline{\includegraphics[width=0.44\textwidth]{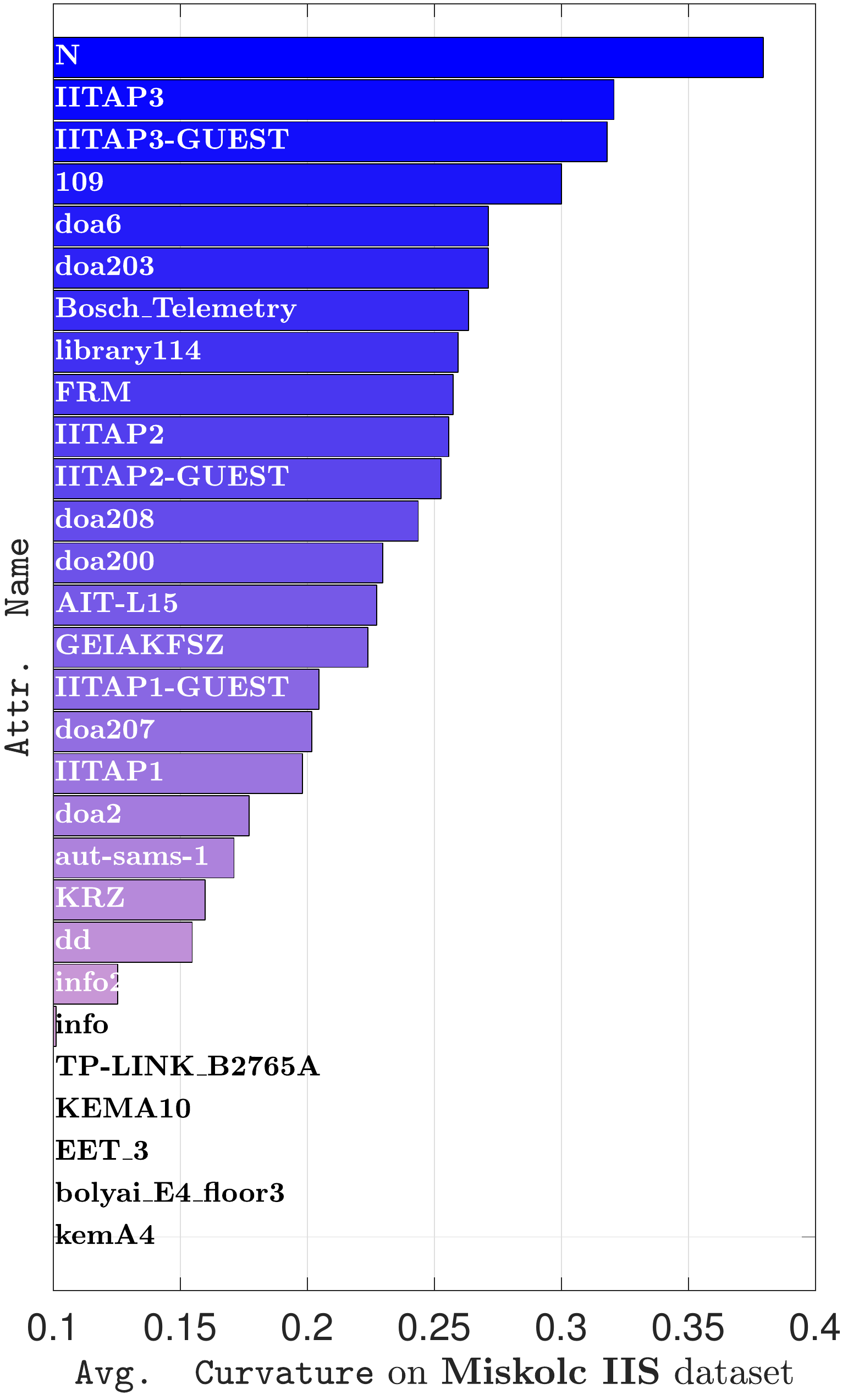}}
	\caption{Feature importance ranking of the Miskolc IIS dataset yieled by CFS.}
	\label{fig:feat_rank}
\end{figure}

To achieve dimensionality reduction, the CFS is adopted\footnote{\textcolor{magenta}{\texttt{\href{https://github.com/zhemingzuo/CFS}{https://github.com/zhemingzuo/CFS}}}} in this study. Furthermore, the feature importance ranking among all the 29 attributes are summarized in Figure \ref{fig:feat_rank}.

\subsection{Experimental Results}
\begin{figure}[!ht]
	\centerline{\includegraphics[width=0.48\textwidth]{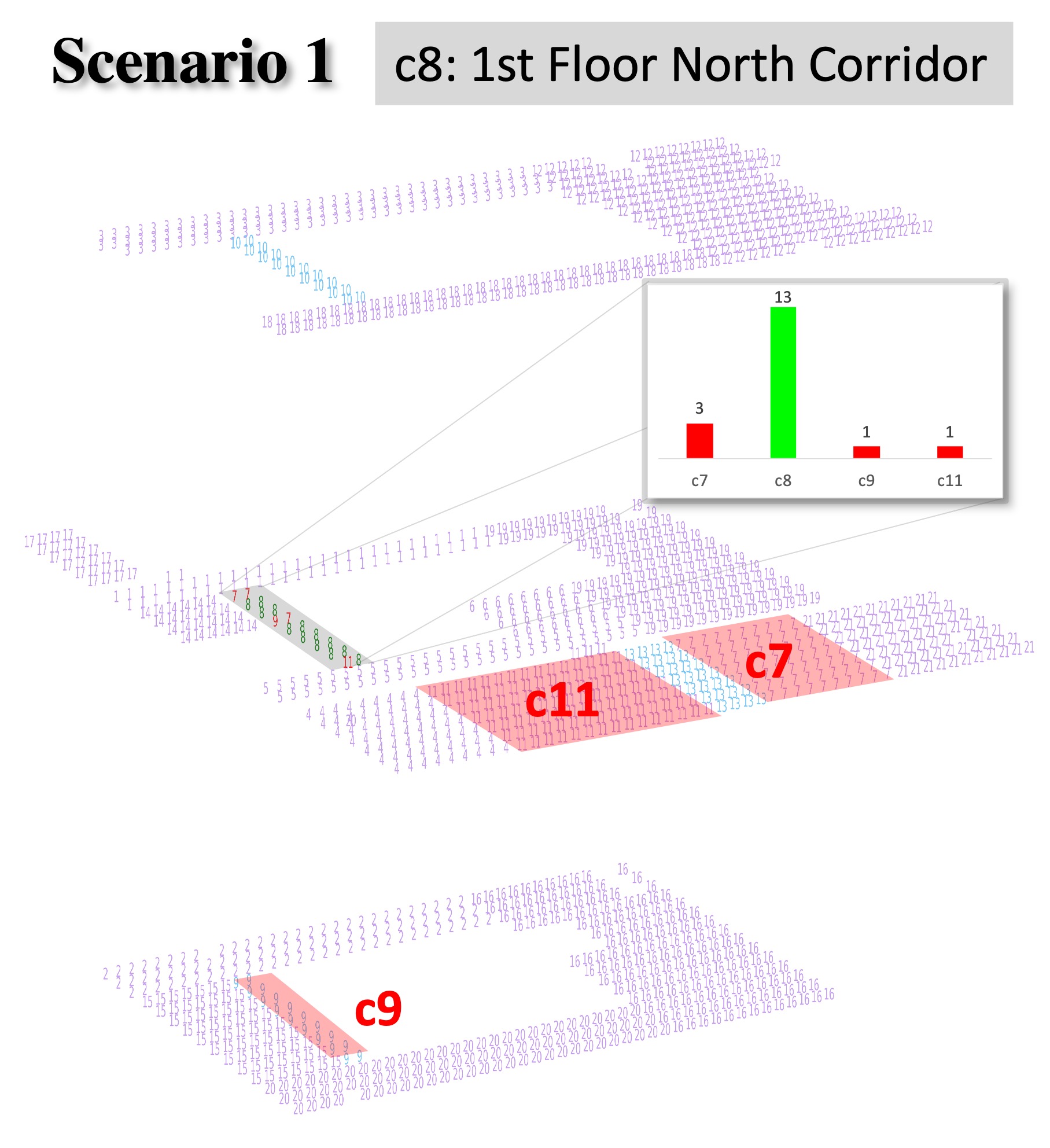}}
	\caption{Experimental results for unseen class label `c8' prediction using CFS.}
	\label{fig:exp_results_w_cfs}
\end{figure}

As described in the aforementioned subsections, we firstly detail the prediction performance using all the 29 attributes (\emph{i.e.} without CFS applied) Figure \ref{fig:exp_results}. This is followed by using feature importance ranking yielded by CFS (see Figure \ref{fig:feat_rank}) to reduce the number of features (from 29 to 8) employed as well as the computational cost for unseen label prediction. It is noteworthy that we only visualize the detailed prediction of our system in conjunction with CFS adopted in Figure \ref{fig:exp_results_w_cfs} due to space limitation. Lastly, all the prediction performances are summarized in Table \ref{tbl:polyu-perf}

One step further, in Figure \ref{fig:exp_results}, it could be drawn the conclusion that the more continuous unseen labels are in the training phase, the more reasonable predictions will be yielded by our system. Specifically, 8, 5 and 9 correct predictions are given by our system in terms of respectively, `c8', `c9' and 'c10', whereas 11 out of the total 28 predictions are correctly given for `c13'. For `c8', we see that all the 10 wrong predictions given by our system are fairly close to the ground-truth `c8', \emph{i.e.} 3, 2 and 5 of which are wrong predicted as `c7', `c9' and `c10'. Similarly, all the 13 wrong predictions were distributed in `c10', `c11' and `c12' when the ground-truth label is `c9'; and all the 9 wrong predictions are covered the labels of `c7', `c8' and `c9' for `c10'. On the contrast, all the 17 wrong prediction are discretely distribute among a wider range of class labels when the unseen label is `c13', \emph{i.e.} `c6', `c7', `c8', `c9', `c11', `c12', `c14' and `c17'.

We adopt CFS from the above findings to use the eight most important features to conduct unseen label prediction using the proposed system. As visualized in Figure \ref{fig:exp_results_w_cfs}, the prediction accuracy of `c8' is 72.22\%. That is, 13 out of a total of 18 predictions are correctly predicted as `c8'. In terms of the rest 5 wrong predictions, three of which are predicted as `c7' (same as that of the ones yielded without using CFS), whereas the rest two are produced as `c9' and `c11'. Furthermore, `c7' (\emph{i.e.} `Lab 101'), `c11' (\emph{i.e.} `Lab 103') and target label `c8' (\emph{i.e.} `1st Floor North Corridor') are all located in the first floor of the building and they are close with each other. In addition, `c9' represents the `2nd Floor North Corridor', and it is possible that the signals provided by IoT devices located in `c8' get interfered from those given by `c9'.

Given the above findings, it is noteworthy that the prediction performance of scenario 1) is generally better than that of 2), which confirms the practical feasibility of our system for the task of unseen label prediction. Additionally, the prediction precision of our system with CFS generally outperforms those without CFS. This, in turn, enables our system to be a reasonable candidate for such a task.


\begin{table}[!ht]
		\setlength{\tabcolsep}{3pt}
		\begin{center}
			\scalebox{0.8}{
				\begin{tabular}{c|c|c|c|c|c}
					
					Scenario & Predicted class & w. CFS & \texttt{Dim.}$\downarrow$ & $\mathcal{D}$ & Accuracy ($\%$)$\uparrow$\\
					\hline
					\multirow{6}{*}{1} & \multirow{2}{*}{c8} & \textcolor{red}{\xmark} & 29 & 8.2 & 44.44\\
				    & & \textcolor{green}{\cmark} & 8 & 9.0 & 72.22\\
				    \cline{2-6}
					& \multirow{2}{*}{c9} & \textcolor{red}{\xmark} & 29 & 3.0 & 27.78\\
					& & \textcolor{green}{\cmark} & 8 & 5.0 & 22.22\\
					\cline{2-6}
					& \multirow{2}{*}{c10} & \textcolor{red}{\xmark} & 29 & 1.0 & 50.00\\
					& & \textcolor{green}{\cmark} & 8 & 1.0 & 50.00\\
					\hline
					\multirow{2}{*}{2} & \multirow{2}{*}{c13} & \textcolor{red}{\xmark} & 29 & 29.0 & 39.29\\
					& & \textcolor{green}{\cmark} & 8 & 21.2 & 53.57\\
					\hline
				\end{tabular}
			}
		\end{center}
		\caption{Quantitative evaluation of our system on the Miskolc IIS dataset.}
		\label{tbl:polyu-perf}
	\end{table}

\section{\uppercase{Conclusions}}\label{sec:conclusion}
A fuzzy inference system has been proposed in this work as an effort to move towards the challenging task of unseen label prediction, which is achieved by adopting the TSK+ interpolation and CFS method. Experimental results on the real-world IoT devices positioning dataset reveal the efficiency and efficacy of the proposed system. Though promising results have been obtained, the involvement of optimization algorithms to improve the generalization capability of the rule base could be an active future direction.

\section*{\uppercase{ACKNOWLEDGEMENT}}
This work is partly supported by VC Research (VCR 0000171) for Prof Chang.
	
	
	\bibliographystyle{unsrt}  
\bibliography{ref}

\begin{thebibliography}{10}

\bibitem{kim2021review}
Tan Kim~Geok, Khaing Zar~Aung, Moe Sandar~Aung, Min Thu~Soe, Azlan Abdaziz,
  Chia Pao~Liew, Ferdous Hossain, Chih~P Tso, and Wong~Hin Yong.
\newblock Review of indoor positioning: Radio wave technology.
\newblock {\em Appl. Sci.}, 11(1):279, 2021.

\bibitem{9500208}
Viachaslau Kachurka, Bastien Rault, Fernando~Ireta Muñoz, David Roussel,
  Fabien Bonardi, Jean-Yves Didier, Hicham Hadj-Abdelkader, Samia Bouchafa,
  Pierre Alliez, and Maxime Robin.
\newblock Weco-slam: Wearable cooperative slam system for real-time indoor
  localization under challenging conditions.
\newblock {\em IEEE Sens. J.}, pages 1--1, 2021.

\bibitem{suroso2021random}
Dwi~Joko Suroso, Alvin~SH Rudianto, Muhammad Arifin, and Singgih Hawibowo.
\newblock Random forest and interpolation techniques for fingerprint-based
  indoor positioning system in un-ideal environment.
\newblock {\em Int. J. Comput. Digital Syst.}, 2021.

\bibitem{9568311}
M.~W.~P. Maduraga and Ruvan Abeysekara.
\newblock Comparison of supervised learning-based indoor localization
  techniques for smart building applications.
\newblock In {\em Proc. IEEE Int. Res. Conf. Smart Comput. Syst. Eng.},
  volume~4, pages 145--148, 2021.

\bibitem{9389926}
Danjue Zhang and Chufan Tan.
\newblock Application of indoor positioning technology in smart home management
  system.
\newblock In {\em Proc. IEEE Int. Conf. Big Data Artif. Intell. Internet Things
  Eng.}, pages 627--631, 2021.

\bibitem{9261582}
Mohamed Abdel-Basset, Hossam Hawash, Victor Chang, Ripon~K. Chakrabortty, and
  Michael Ryan.
\newblock Deep learning for heterogeneous human activity recognition in complex
  iot applications.
\newblock {\em IEEE Internet Things J}, pages 1--1, 2020.

\bibitem{vcchang_iot_pos}
Victor Chang, Yeqing Mou, and Qianwen~Ariel Xu.
\newblock The ethical issues of location-based services on big data and iot.
\newblock In {\em Proc. Int. Conf. Ind. IoT Big Data Supply Chain}, pages
  195--205, 2021.

\bibitem{khanh2020wi}
Tran~Trong Khanh, VanDung Nguyen, Xuan-Qui Pham, and Eui-Nam Huh.
\newblock Wi-fi indoor positioning and navigation: a cloudlet-based cloud
  computing approach.
\newblock {\em Human-centric Comput. Info. Sci.}, 10(1):1--26, 2020.

\bibitem{curran2011evaluation}
Kevin Curran, Eoghan Furey, Tom Lunney, Jose Santos, Derek Woods, and Aiden
  McCaughey.
\newblock An evaluation of indoor location determination technologies.
\newblock {\em J. Location Based Serv.}, 5(2):61--78, 2011.

\bibitem{jordan2015machine}
Michael~I Jordan and Tom~M Mitchell.
\newblock Machine learning: Trends, perspectives, and prospects.
\newblock {\em Science}, 349(6245):255--260, 2015.

\bibitem{8552138}
Pranesh Sthapit, Hui-Seon Gang, and Jae-Young Pyun.
\newblock Bluetooth based indoor positioning using machine learning algorithms.
\newblock In {\em Proc. IEEE Int. Conf. Consum. Electron. - Asia}, pages
  206--212, 2018.

\bibitem{lecun2015deep}
Yann LeCun, Yoshua Bengio, and Geoffrey Hinton.
\newblock Deep learning.
\newblock {\em nature}, 521(7553):436--444, 2015.

\bibitem{8919994}
Z.~{Zuo}, L.~{Yang}, Y.~{Liu}, F.~{Chao}, R.~{Song}, and Y.~{Qu}.
\newblock Histogram of fuzzy local spatio-temporal descriptors for video action
  recognition.
\newblock {\em IEEE Trans. Ind. Inf.}, 16(6):4059--4067, 2020.

\bibitem{9049390}
W.~{Yang}, Y.~{Yuan}, W.~{Ren}, J.~{Liu}, W.~J. {Scheirer}, Z.~{Wang},
  T.~{Zhang}, et~al.
\newblock Advancing image understanding in poor visibility environments: A
  collective benchmark study.
\newblock {\em IEEE Trans. Image Process.}, 29:5737--5752, 2020.

\bibitem{Zuo_IdeaNet_2022_WACV}
Z.~Zuo, X.~Chen, H.~Xu, J.~Li, W.~Liao, Z.-X. Yang, and S.~Wang.
\newblock Idea-net: Adaptive dual self-attention network for single image
  denoising.
\newblock In {\em Proc. IEEE/CVF Winter Conf. Appl. of Comput. Vis. Work.},
  pages 739--748, January 2022.

\bibitem{nessa2020survey}
Ahasanun Nessa, Bhagawat Adhikari, Fatima Hussain, and Xavier~N Fernando.
\newblock A survey of machine learning for indoor positioning.
\newblock {\em IEEE Access}, 8:214945--214965, 2020.

\bibitem{YangZuoInTech2017}
Longzhi Yang, Zheming Zuo, Fei Chao, and Yanpeng Qu.
\newblock Fuzzy interpolation systems and applications.
\newblock In S.~Ramakrishnan, editor, {\em Modern Fuzzy Control Systems and Its
  Applications}. IntechOpen, Rijeka, 2017.

\bibitem{mamdani1977application}
Ebrahim~H Mamdani.
\newblock Application of fuzzy logic to approximate reasoning using linguistic
  synthesis.
\newblock {\em IEEE Trans. Comput.}, 26(12):1182--1191, 1977.

\bibitem{TSK1985}
T.~Takagi and M.~Sugeno.
\newblock Fuzzy identification of systems and its applications to modeling and
  control.
\newblock {\em ” IEEE Trans. Syst., Man, and Cybern.}, 15(1):116--132, 1985.

\bibitem{koczy1993approximate}
L{\'a}szl{\'o}T K{\'o}czy and Kaoru Hirota.
\newblock Approximate reasoning by linear rule interpolation and general
  approximation.
\newblock {\em Int. J. Approximate Reasoning}, 9(3):197--225, 1993.

\bibitem{Jie2017}
J.~Li, Y.~Qu, H.~P.~H. Shum, and L.~Yang.
\newblock Tsk inference with sparse rule bases.
\newblock In {\em Proc. Springer Adv. Comput. Intell. Syst.}, pages 107--123,
  2017.

\bibitem{li2018interval}
Jie Li, Longzhi Yang, Xin Fu, Fei Chao, and Yanpeng Qu.
\newblock Interval type-2 tsk+ fuzzy inference system.
\newblock In {\em Proc. IEEE Int. Conf. Fuzzy Syst.}, pages 1--8, 2018.

\bibitem{li2018extended}
J~Li, L~Yang, Y~Qu, and G~Sexton.
\newblock An extended takagi--sugeno--kang inference system (tsk+) with fuzzy
  interpolation and its rule base generation.
\newblock {\em Soft Comput.}, 22(10):3155--3170, 2018.

\bibitem{li2016experience}
Jie Li, Hubert~PH Shum, Xin Fu, Graham Sexton, and Longzhi Yang.
\newblock Experience-based rule base generation and adaptation for fuzzy
  interpolation.
\newblock In {\em Proc. IEEE Int. Conf. Fuzzy Syst.}, pages 102--109, 2016.

\bibitem{zuo2021curvature}
Zheming Zuo, Jie Li, Han Xu, and Noura~Al Moubayed.
\newblock Curvature-based feature selection with application in classifying
  electronic health records.
\newblock {\em Elsevier Technol. Forecasting Social Change}, 173:121--127,
  2021.

\bibitem{leger1999menger}
J.-C. L{\'e}ger.
\newblock Menger curvature and rectifiability.
\newblock {\em Ann. Math.}, 149:831--869, 1999.

\bibitem{zuo2018grooming}
Z.~Zuo, J.~Li, P.~Anderson, L.~Yang, and N.~Naik.
\newblock Grooming detection using fuzzy-rough feature selection and text
  classification.
\newblock In {\em Proc. IEEE Int. Conf. Fuzzy Syst.}, pages 1--8, 2018.

\bibitem{8858838}
Z.~{Zuo}, J.~{Li}, B.~{Wei}, L.~{Yang}, F.~{Chao}, and N.~{Naik}.
\newblock Adaptive activation function generation for artificial neural
  networks through fuzzy inference with application in grooming text
  categorisation.
\newblock In {\em Proc. IEEE Int. Conf. Fuzzy Syst.}, pages 1--6, 2019.

\bibitem{kodinariya2013review}
Trupti~M Kodinariya and Prashant~R Makwana.
\newblock Review on determining number of cluster in k-means clustering.
\newblock {\em Int. J. Adv. Res. Comput. Sci. Manage. Stud.}, 1(6):90--95,
  2013.

\bibitem{9138275}
Matteo D'Aloia, Annalisa Longo, Gianluca Guadagno, Mariano Pulpito, Paolo
  Fornarelli, Pietro~Nicola Laera, Dario Manni, and Maria Rizzi.
\newblock Iot indoor localization with ai technique.
\newblock In {\em Proc. IEEE Int. Work. Metrol. Ind. 4.0 IoT}, pages 654--658,
  2020.

\bibitem{alhajri2018classification}
Mohamed~I AlHajri, Nazar~T Ali, and Raed~M Shubair.
\newblock Classification of indoor environments for iot applications: A machine
  learning approach.
\newblock {\em IEEE Antennas Wirel. Propag. Lett.}, 17(12):2164--2168, 2018.

\bibitem{toth2016miskolcIIS}
Judit~Tamás Zsolt~Tóth.
\newblock Miskolc iis hybrid ips: Dataset for hybrid indoor positioning.
\newblock In {\em Proc. IEEE Int. Conf. Radioelektronika}, pages 408--412,
  2016.

\end{thebibliography}

\end{document}